\documentclass[letterpaper, 10 pt, journal, twoside]{ieeetran} % Use this command for final RAL version

\IEEEoverridecommandlockouts                              % This command is only needed if 
\usepackage{amsfonts}	% Additional math fonts
\usepackage{amsmath}	% Math symbols
\usepackage{amssymb}    % Extended math symbols
\usepackage{siunitx}
\usepackage{mathtools}

\usepackage{cuted} % supplement 
\usepackage{placeins}  % supplement

\usepackage{booktabs}
\usepackage{makecell}  % Line breaks in table cells
\usepackage[flushleft]{threeparttable}  % For notes below a table
\usepackage{multirow}

\usepackage{xspace}    % for correct spacing after defined commands without arguments, e.g., \net
\usepackage[table]{xcolor}
\usepackage{xcolor}    % for colored text
\usepackage{textcomp}  % prevent warnings of gensymb package
\usepackage{gensymb}   % \degree
\usepackage{graphicx} % insert PDF as figure

\makeatletter
\let\NAT@parse\undefined
\makeatother

\usepackage[pdfencoding=auto, hidelinks]{hyperref} % Usually, this is the last package to be imported
\usepackage[hang,flushmargin]{footmisc}

\usepackage[capitalize]{cleveref}
\crefname{section}{Sec.}{Secs.}
\Crefname{section}{Section}{Sections}
\Crefname{table}{Table}{Tables}
\crefname{table}{Tab.}{Tabs.}

\DeclareMathOperator*{\argmax}{arg\,max}
\DeclareMathOperator*{\argmin}{arg\,min}
\DeclareMathOperator*{\atantwo}{atan2}

\definecolor{Gray}{gray}{0.9}
\newcommand{\grayrule}{\arrayrulecolor{black!30}\midrule\arrayrulecolor{black}}

\newcommand{\net}{\mbox{MDPCalib}\xspace}

\renewcommand{\thefootnote}{\fnsymbol{footnote}}

\begin{document}

\title{Automatic Target-Less Camera-LiDAR Calibration From Motion and Deep Point Correspondences}

\author{
Kürsat Petek$^{1*}$,
Niclas Vödisch$^{1*}$,
Johannes Meyer$^{1}$,
Daniele Cattaneo$^{1}$,
Abhinav Valada$^{1}$,
Wolfram Burgard$^{2}$% <-this % stops a space
\thanks{$^{*}$ Equal contribution.}%
\thanks{© 2024 IEEE. Personal use of this material is permitted. Permission from IEEE must be obtained for all other uses, in any current or future media, including reprinting/republishing this material for advertising or promotional purposes, creating new collective works, for resale or redistribution to servers or lists, or reuse of any copyrighted component of this work in other works.}
\thanks{This work was funded by the German Research Foundation Emmy Noether Program grant No 468878300 and an academic grant from NVIDIA.}%
\thanks{$^{1}$ Kürsat Petek, Niclas Vödisch, Johannes Meyer, Daniele Cattaneo, and Abhinav Valada are with the Department of Computer Science, University of Freiburg, Germany.}%
\thanks{$^{2}$ Wolfram Burgard is with the Department of Engineering, University of Technology Nuremberg, Germany.}%
\thanks{Digital Object Identifier (DOI): 10.1109/LRA.2024.3468090}
}

% Paper headers
\markboth{T\MakeLowercase{his paper appeared in:} IEEE ROBOTICS AND AUTOMATION LETTERS, VOL. 9, ISSUE 11, NOVEMBER 2024}%
{Petek \MakeLowercase{\textit{et al.}}: Automatic Target-Less Camera-LiDAR Calibration From Motion and Deep Point Correspondences}
% Use only for final RAL version

\maketitle

%%%%%%%%%%%%%%%%%%%%%%%%%%%%%%%%%%%%%%%%%%%%%%%%%%%%%%%%%%%%%%%%%%%%%%%%%%%%%%%%

\begin{abstract}
    Sensor setups of robotic platforms commonly include both camera and LiDAR as they provide complementary information. However, fusing these two modalities typically requires a highly accurate calibration between them. In this paper, we propose \net which is a novel method for camera-LiDAR calibration that requires neither human supervision nor any specific target objects. Instead, we utilize sensor motion estimates from visual and LiDAR odometry as well as deep learning-based 2D-pixel-to-3D-point correspondences that are obtained without in-domain retraining. We represent camera-LiDAR calibration as an optimization problem and minimize the costs induced by constraints from sensor motion and point correspondences. In extensive experiments, we demonstrate that our approach yields highly accurate extrinsic calibration parameters and is robust to random initialization. Additionally, our approach generalizes to a wide range of sensor setups, which we demonstrate by employing it on various robotic platforms including a self-driving perception car, a quadruped robot, and a UAV.
To make our calibration method publicly accessible, we release the code on our project website at \mbox{\url{http://calibration.cs.uni-freiburg.de}}.

\end{abstract}

%%%%%%%%%%%%%%%%%%%%%%%%%%%%%%%%%%%%%%%%%%%%%%%%%%%%%%%%%%%%%%%%%%%%%%%%%%%%%%%%


%%%%%%%%%%%%%%%%%%%%%%%%%%%%%%%%%%%%%%%%%%%%%%%%%%%%%%%%%%%%%%%%%%%%%%%%%%%%%%%%

\section{Introduction}

\IEEEPARstart{S}ensor fusion for robotic systems has been extensively investigated~\cite{andresen2020accurate, liang2022bevfusion} as it promises to efficiently combine complementary information from different modalities, e.g., to increase robustness in case of sensor failures~\cite{ge2023metabev} and towards weather conditions~\cite{schramm2023bevcar}. However, the effectiveness of fusion approaches depends heavily on the extrinsic calibration between the sensors such as cameras and LiDAR.

Due to the importance of the task, camera-LiDAR calibration has been widely studied by the research community. Previously proposed methods can generally be classified into target-based and target-less approaches. Approaches of the first category often rely on artificial patterns such as checkerboards~\cite{dhall2017lidar, zhang2004extrinsic} and require manual labor or another kind of human supervision to associate 2D points from the image space with 3D points in the LiDAR point cloud~\cite{gtcalibration}.
While a substantial effort has gone into the detection of the target and the automation of the matching process~\cite{guindel2017automatic, kim2019extrinsic}, calibration often still needs special data collection.
Some target-less calibration methods intend to overcome this problem, e.g., by inferring the extrinsic transform from the sensor motion~\cite{zhang2023an} or by matching vision-based structure-from-motion models with accumulated point clouds from the LiDAR~\cite{tu2022multicamera}. Although these approaches are generally more widely applicable as they enable sensor calibration from normal robot operation, they often still require an initial set of parameters.

\begin{figure}[t]
    \centering
    \includegraphics[width=.9\linewidth]{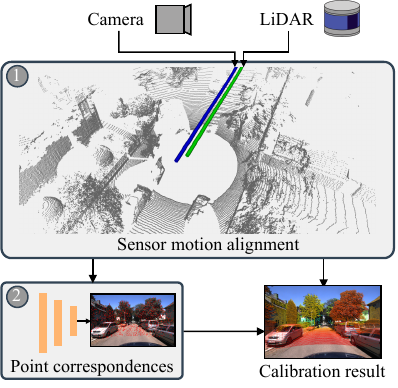}
    \vspace{-.3cm}
    \caption{Our proposed method, called \net, for camera-LiDAR calibration comprises two steps: We first initialize the extrinsic parameters by aligning the motion of both sensors. Afterward, we refine the calibration results by leveraging deep learning-based 2D-to-3D point correspondences.}
    \label{fig:cover}
    \vspace{-.3cm}
\end{figure}

In this work, we propose the novel \net to eliminate this drawback by fully automating the calibration procedure. Given recorded data from normal robot operation, we employ visual and LiDAR odometry to generate two paths that can be aligned via non-linear optimization for coarse sensor registration. Afterward, we use the coarse parameters to initialize a learning-based 2D-to-3D point correspondence algorithm that outputs dense matches between the image and the point cloud spaces. In the final step, we jointly optimize with respect to both sensor motion and point correspondence, thereby effectively fusing the complementary information. Robust loss functions account for outlying observations from either source.
In summary, the main contributions are:
\begin{enumerate}
    \item We introduce \net for automatic target-less camera-LiDAR calibration that requires neither human initialization nor special data recording.
    \item We propose to formulate extrinsic calibration as an optimization problem constrained by sensor motion and deep learning-based point correspondences.
    \item We extensively demonstrate the general applicability of \net to both public and in-house datasets.
    \item We release our code along with a detailed user guide on \mbox{\url{http://calibration.cs.uni-freiburg.de}}.
\end{enumerate}

\section{Related Work}

Methods for extrinsic camera-LiDAR calibration can generally be categorized into target-based and target-less approaches. While the former have been well investigated for several years, recent advances in deep learning have enabled the rise of target-less calibration removing the need for explicit target objects. In this section, we present an overview of approaches from both categories.

%%%%%%%%%%%%%%%%%%%%%%%%%%%%%%%%%%%%%%%%%%%%%%%%%%%%%%%%%%%%%%%%%%%%%%%%%%%%%%%%

{\parskip=3pt
\noindent\textit{Target-Based Calibration:} 
Inspired by the estimation of camera parameters with a checkerboard pattern~\cite{zhang1999flexible}, Zhang~\textit{et~al.}~\cite{zhang2004extrinsic} were the first to propose using a similar target also for extrinsic calibration between a camera and a LiDAR.
Since then, many different styles of patterns have been described to further optimize this calibration procedure both in terms of accuracy and applicability. For instance, Dhall~\textit{et~al.}~\cite{dhall2017lidar} exploit ArUco markers with known sizes to get accurate estimates of the corner points of the pattern in 3D, which are then registered to the corner point detected by the LiDAR using the ICP algorithm. Similarly, Kim~\textit{et~al.}~\cite{kim2019extrinsic} fit points on a checkerboard detected by a camera to the corresponding plane in the LiDAR point cloud.
However, not all target-based approaches rely on a checkerboard pattern or its variants to establish point correspondences. For example, both Velas~\textit{et~al.}~\cite{velas2014calibration} and Guindel~\textit{et~al.}~\cite{guindel2017automatic} utilize wooden boards with holes to obtain point correspondences.
}

Similar to our method, Ou~\textit{et~al.}~\cite{ou2022automatic} propose to formulate camera-LiDAR registration as a graph optimization problem. In particular, their method first extracts corner points in both modalities to perform an initial calibration obtained by a perspective-n-point (PnP) algorithm. This allows reprojecting the LiDAR points onto the image plane and computing the reprojection error, which is then used as a cost term in the graph-based formulation that can be efficiently solved using graph optimization methods~\cite{kuemmerle2011g2o}.

%%%%%%%%%%%%%%%%%%%%%%%%%%%%%%%%%%%%%%%%%%%%%%%%%%%%%%%%%%%%%%%%%%%%%%%%%%%%%%%%

{\parskip=3pt
\noindent\textit{Target-Less Calibration:}
Target-less calibration aims at performing camera-LiDAR registration without specifically designed target objects. For robotics, this opens an avenue for flexible and potentially online recalibration and enables applications to large fleets by reducing human supervision.
}

Correspondence-based methods replace the artificial targets with patterns that can be perceived in structured environments such as urban areas. For instance, Yuan~\textit{et~al.}~\cite{yuan2021pixel} and Yin~\textit{et~al.}~\cite{yin2023automatic} match edges obtained from both images and LiDAR point clouds. 
Tu~\textit{et~al.}~\cite{tu2022multicamera} extract features of structure-from-motion (SfM) points of camera data and LiDAR points, followed by optimizing these correspondences jointly with camera intrinsics as well as camera and LiDAR poses. Koide~\textit{et~al.}~\cite{koide2023general} build on the commonly used normalized information distance that poses a distance metric between the image and projected LiDAR points to measure the amount of mutual information. Finally, Caselitz~\textit{et~al.}~\cite{caselitz2016monocular} propose a method for determining the pose of an RGB camera with respect to a 3D point cloud generated from LiDAR data by matching geometric clues. 
On the other hand, correspondence-free methods often rely on leveraging output data of auxiliary tasks such as monocular depth prediction~\cite{zhu2023calibdepth, borer2024chaos} or sensor motion estimation~\cite{zhang2023an, yin2023automatic, taylor2015motion}.
Both Zhang~\textit{et~al.}~\cite{zhang2023an} and Yin~\textit{et~al.}~\cite{yin2023automatic} match trajectories from visual and LiDAR odometry and obtain extrinsic parameters via optimization. The latter then utilize these parameters to initialize an edge-driven refinement stage.
Finally, a more direct approach is enabled by exploiting deep learning-based correspondences between RGB images and LiDAR point clouds. Both RegNet~\cite{schneider2017regnet} and CMRNext~\cite{cattaneo2024cmrnext} involve training multiple CNNs on varying levels of decalibration and employ these networks during test time in a hierarchical manner. In detail, CMRNext frames the point-to-pixel matching problem as an optical flow estimation task. Building upon this approach, LCCNet~\cite{lv2021lccnet} proposes the construction of a cost volume that stores matching costs. However, as it predicts the 6-DoF rigid-body transformation, it suffers from a high dependency on the training setup and is thus less generalizable.

In our proposed method, we combine the advantages of both correspondence-free and correspondence-based approaches by performing coarse initialization based on sensor motion followed by fine registration incorporating deep learning-based point correspondences~\cite{cattaneo2024cmrnext}. Particularly, in contrast to previous works~\cite{yin2023automatic}, we utilize robust cost functions and jointly optimize with respect to both sensor motion and point correspondences, accounting for their complimentary information and increasing robustness towards outlying observations.

\section{Technical Approach}

\begin{figure*}[t]
    \centering
    \includegraphics[width=\linewidth]{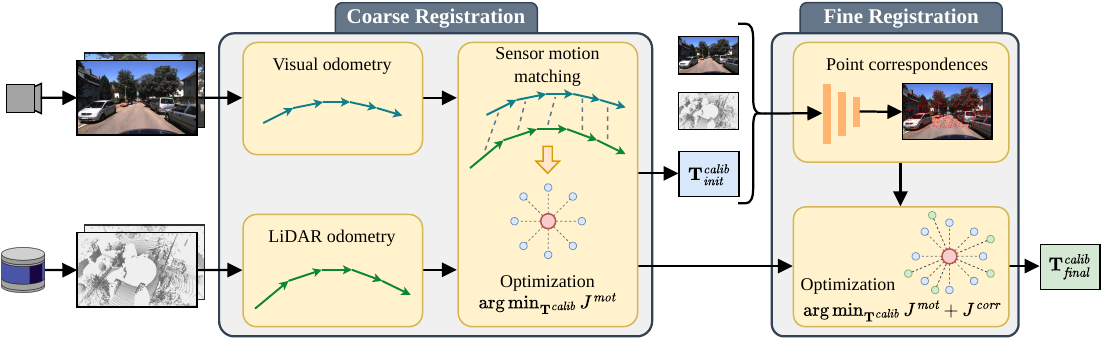}
    \vspace*{-.6cm}
    \caption{Our proposed method for camera-LiDAR calibration processes two input streams of RGB images and 3D point clouds. The first step comprises a coarse registration based on sensor motion estimated with visual and LiDAR odometry. These motion estimates yield time-synchronized matches serving as constraints in an optimization problem. Given the obtained initial calibration parameters, a neural network is used to find 2D pixel to 3D point correspondences that result in additional constraints. The second step consists of joint optimization with respect to both sensor motion and point correspondences yielding the overall extrinsic calibration parameters.}
    \label{fig:overview}
    \vspace*{-.5cm}
\end{figure*}

In this section, we present our \net approach for automatic target-less camera-LiDAR calibration.
As illustrated in \cref{fig:overview}, \net comprises two consecutive steps by combining a coarse initialization from sensor motion with a fine-tuning stage that takes learning-based correspondences between image pixels and \mbox{LiDAR} points into account. We first provide the relevant mathematical background, then introduce the general problem formulation, and finally give a detailed description of both registration steps.

%%%%%%%%%%%%%%%%%%%%%%%%%%%%%%%%%%%%%%%%%%%%%%%%%%%%%%%%%%%%%%%%%%%%%%%%%%%%%%%%

\subsection{Mathematical Preliminaries}

We interpret camera-LiDAR calibration as an optimization problem that determines the most likely transformation between the coordinate frames of a camera and a LiDAR given a set of sensor measurements. Mathematically, this can be represented by a conditional probability distribution:
\begin{equation}
    P( x \mid x_0, z_{0:k} ) \, ,
    \label{eqn:cond-prob}
\end{equation}
where $x$ denotes the state vector with an initial guess $x_0$ and $z_{0:k}$ refers to a set of observations, i.e., sensor data. Instead of calculating the exact probability distribution, we perform maximum a posteriori (MAP) estimation assuming Gaussian distributions and independent and identically distributed measurements. This yields the optimal state~$x^*$:
\begin{equation}
    x^* = \argmax_x P( x \mid x_0, z_{0:k} ) \, .
\end{equation}
In practice, we solve for $x^*$ by optimizing a robustified non-linear squares problem of the form:
\begin{subequations}
\begin{align}
    x^* &= \argmin_x \sum\nolimits_i \rho_i \bigl( \lVert f_i (x, z_i) \rVert_2^2 \bigr) \\
        &= \argmin_x \sum\nolimits_i \rho_i \bigl( e_i(x, z_i)^T \mathbf{\Omega}_i e_i(x, z_i) \bigr) \, ,
\end{align}
\end{subequations}
where $\rho_i$ denotes the robustifier and $f_i$ refers to a cost function applied to observation~$z_i$. For efficiency, we compute the squared Frobenius norm $\lVert \cdot \rVert_2^2$ via vector multiplication of the induced error vectors $e_i \in \mathbb{R}^{d \times 1}$ with a diagonal weighting matrix $\mathbf{\Omega}_i \in \mathbb{R}^{d \times d}$. We utilize the Cauchy loss as robustifier due to its high tolerance to outliers in the observations~\cite{liu2014cauchy}, resulting in a significant increase in calibration accuracy.

To obtain the MAP estimate~$x^*$, we leverage an optimization formulation~\cite{kuemmerle2011g2o} with a single state~$x$ and multiple constraints defined by the error terms~$e_i$. In particular, we define the state~$x$ as the calibration parameters~$\mathbf{T}^\mathit{calib}$ and decompose the sum into three error types:
\begin{equation}
    \mathbf{T}^\mathit{calib}_\mathit{final} = \argmin_{\mathbf{T}^\mathit{calib}} \underbrace{\sum_{i=0}^{k-1} \dots}_{J^\mathit{rot}} + \underbrace{\sum_{i=k}^{2k-1} \dots}_{J^\mathit{trans}} + \underbrace{\sum_{i=2k}^{2k+m} \dots}_{J^\mathit{corr}} \, ,
    \label{eqn:error-terms}
\end{equation}
with $k$ observations for sensor motion with rotational and translational costs $J^\mathit{rot}$ and $J^\mathit{trans}$ and $m$ observations for the point correspondences cost $J^\mathit{corr}$.

%%%%%%%%%%%%%%%%%%%%%%%%%%%%%%%%%%%%%%%%%%%%%%%%%%%%%%%%%%%%%%%%%%%%%%%%%%%%%%%%

\subsection{Coarse Registration}
\label{ssec:coarse-registration}

During the first step of \net, we perform coarse camera-LiDAR calibration by matching sensor motion. In particular, we use both vision- and LiDAR-based odometry to obtain poses $\mathbf{P}^\mathit{cam}_{0:k}$ and $\mathbf{P}^\mathit{lidar}_{0:k}$ capturing oriented positions of the camera and the LiDAR, respectively.
To estimate the camera poses, we employ ORB-SLAM3~\cite{campos2021orbslam}. In contrast to deep learning-based methods~\cite{voedisch2023covio, zhan2019dfvo}, classical feature-based tracking approaches such as ORB-SLAM3 are more robust towards detecting lost tracks, which would lead to inconsistent constraints in the optimization problem.
For the LiDAR, we perform consecutive scan matching using an adapted version of \mbox{FAST-LIO2}~\cite{xu2022fastlio2} without the measurements of an inertial measurement unit (IMU). If not already done during the post-processing of the sensor output, the LiDAR scans are undistorted before being matched.
Given time-synchronized sensor measurements, we now interpolate the poses of the LiDAR such that we obtain pose pairs of both sensors at each image timestamp. As detailed in \cref{ssec:pose-interp}, we further transform the nearest point cloud to the same timestamp using the estimated LiDAR odometry as this data will be processed during the fine-tuning stage. Subsequently, we compute the pose difference between two consecutive poses yielding homogeneous transforms $\mathbf{T}_i^\mathit{sensor}$ with $\mathit{sensor} \in \{\mathit{cam}, \mathit{lidar}\}$. Using these transforms, we solve the following equation to find the extrinsic calibration parameters $\mathbf{T}^\mathit{calib}$:
\begin{equation}
    \mathbf{T}^\mathit{cam} \mathbf{T}^\mathit{calib} = \mathbf{T}^\mathit{calib} \mathbf{T}^\mathit{lidar} \, .
    \label{eqn:hand-eye}
\end{equation}

As derived by Shiu~\textit{et~al.}~\cite{shiu1989calibration}, such an equation can be decomposed into solving for the rotational and translational components separately:
\begin{subequations}
\label{eqn:hand-eye-decomp}
\begin{align}
    \mathbf{R}^\mathit{cam} \mathbf{R}^\mathit{calib} &= \mathbf{R}^\mathit{calib} \mathbf{R}^\mathit{lidar} \, ,
    \label{eqn:hand-eye-rot} \\
    \mathbf{R}^\mathit{cam} \mathbf{t}^\mathit{calib} + s \, \mathbf{t}^\mathit{cam} &= \mathbf{R}^\mathit{calib} \mathbf{t}^\mathit{lidar} + \mathbf{t}^\mathit{calib} \, .
    \label{eqn:hand-eye-trans}
\end{align}
\end{subequations}
$\mathbf{R} \in \mathbb{R}^{3 \times 3}$ denotes a rotation matrix and $\mathbf{t} \in \mathbb{R}^{3 \times 1}$ is a translation vector. Since ORB-SLAM3 does not produce metrically aware odometry estimates, we add a scaling factor~$s$.
Next, we apply \cref{eqn:hand-eye} to all paired pose differences $\left( \mathbf{T}_i^\mathit{cam}, \mathbf{T}_i^\mathit{lidar} \right)$ with $i = [1, k]$ and define $J^\mathit{rot}$ and $J^\mathit{trans}$ as follows:
\begin{subequations}
\begin{align}
    J^\mathit{rot} &= \sum\nolimits_i \rho_i \bigl( e_{i, \mathit{rot}}^T \mathbf{\Omega}_\mathit{rot} e_{i, \mathit{rot}} \bigr) \, , \\
    J^\mathit{trans} &= \sum\nolimits_i \rho_i \bigl( e_{i, \mathit{trans}}^T \mathbf{\Omega}_\mathit{trans} e_{i, \mathit{trans}} \bigr) \, ,
\end{align}
\end{subequations}
with error functions $e_\mathit{rot}$ and $e_\mathit{trans}$ induced by \cref{eqn:hand-eye-rot} and \cref{eqn:hand-eye-trans}, respectively:
\begin{subequations}
\begin{align}
    \mathbf{e}_{i, \mathit{rot}} &= \left[ \left( \mathbf{R}^\mathit{calib} \mathbf{R}^\mathit{lidar}_i \right)^{-1} \left( \mathbf{R}^\mathit{cam}_i \mathbf{R}^\mathit{calib} \right) - \mathbf{I} \right]_\mathit{vec} \, , \\
    e_{i, \mathit{trans}} &= \left( \mathbf{R}^\mathit{cam}_i - \mathbf{I} \right) \mathbf{t}^\mathit{calib} + s_i \, \mathbf{t}^\mathit{cam}_i - \mathbf{R}^\mathit{calib} \mathbf{t}^\mathit{lidar}_i \, ,
\end{align}
\end{subequations}
where $\mathbf{I}$ denotes the identity matrix. We further define the operator $[\cdot]_\mathit{vec}$ that reshapes a matrix $\mathbf{M} \in \mathbb{R}^{d \times d}$ to $\mathbb{R}^{d^2 \times 1}$ by stacking its columns.
Note that we use the same information matrices $\mathbf{\Omega}_\mathit{rot}$ and $\mathbf{\Omega}_\mathit{trans}$ for all data pairs.
In \cref{fig:interpolation}, we refer to the tuple of both error functions as \textit{odometry constraints}.
Finally, we perform MAP estimation using the combined sensor motion cost $J^\mathit{mot}$ to obtain the initial calibration parameters $\mathbf{T}^\mathit{calib}_\mathit{init}$ and scaling factors $s_{1:k}$\footnote{We omit these in the optimization equations to improve readability.}:
\begin{equation}
    \mathbf{T}^\mathit{calib}_\mathit{init} = \argmin_{ \mathbf{T}^\mathit{calib}} J^\mathit{rot} + J^\mathit{trans} = \argmin_{ \mathbf{T}^\mathit{calib}} J^\mathit{mot} \, .
\end{equation}

%%%%%%%%%%%%%%%%%%%%%%%%%%%%%%%%%%%%%%%%%%%%%%%%%%%%%%%%%%%%%%%%%%%%%%%%%%%%%%%%

\begin{figure}
    \centering
    \includegraphics[width=\linewidth]{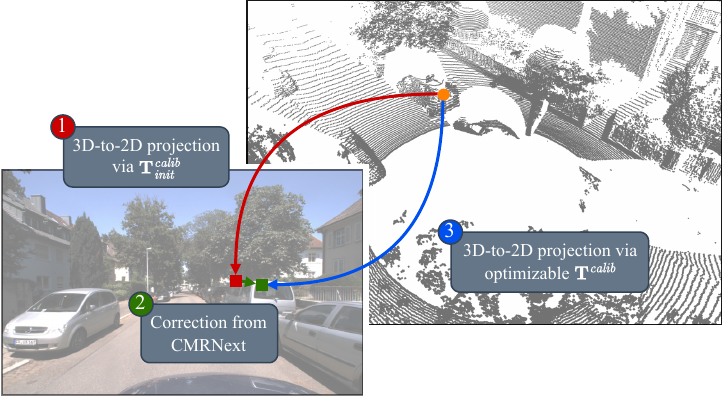}
    \vspace{-.6cm}
    \caption{In the fine-tuning stage, we employ \mbox{CMRNext}~\cite{cattaneo2024cmrnext} to find 2D pixel to 3D point correspondences. First, a LiDAR point is projected onto the image space using the coarse calibration parameters. Second, CMRNext predicts a 2D offset to correct the projection. Finally, during optimization, the calibration parameters are adjusted to match the corrected projection.}
    \label{fig:cmrnext-constraint}
    \vspace{-.5cm}
\end{figure}

\subsection{Fine Registration}
\label{ssec:fine-registration}

The second step of \net extends the coarse registration from sensor motion with matching correspondences between image pixels and LiDAR points. 
In contrast to prior works~\cite{yin2023automatic, taylor2015motion}, we propose to perform joint optimization with respect to both constraints explicitly exploiting their complementary information. We demonstrate the superiority of this design in \cref{ssec:ablation-studies}.
As shown in \cref{fig:overview}, we construct input triplets comprising an image $\mathit{Img}^\mathit{cam}_j$, a synchronized point cloud $\mathit{Pcl}^\mathit{lidar}_j$ (see \cref{ssec:pose-interp}), and the calibration parameters $\mathbf{T}^\mathit{calib}_\mathit{init}$ from coarse registration. This data is fed to the deep learning-based CMRNext~\cite{cattaneo2024cmrnext} that is able to register a camera frame to a 3D point cloud. We select CMRNext due to its generalizability to new scenes and sensor models. We illustrate the detailed steps in \cref{fig:cmrnext-constraint}: First, a 3D point is projected to a 2D pixel coordinate based on the provided initial guess. Second, CMRNext estimates an offset to correct the initial guess. This enhancement step is performed iteratively with network weights trained for decreasing offsets. Formally,
\begin{equation}
    p^\mathit{cmr}_j = \mathit{CMRNext} \left( p^\mathit{lidar}_j, \mathit{Img}^\mathit{cam}_j, \mathbf{T}^\mathit{calib}_\mathit{init} \right) \, ,
\end{equation}
with point $p^\mathit{lidar}_j \in \mathit{Pcl}^\mathit{lidar}_j$.
Third, we define a error function~$e_\mathit{corr}$ that attempts to gradually alter the calibration parameters $\mathbf{T}^\mathit{calib}$ such that the direct 3D-to-2D projection approaches the estimate of CMRNext:
\begin{equation}
    e_{j, \mathit{corr}} = \mathit{proj}( p^\mathit{lidar}_j, \mathbf{K}, \mathbf{T}^\mathit{calib} ) - p^\mathit{cmr}_j \, ,
\end{equation}
where $\mathbf{K}$ denotes the camera matrix.
This error function, referred to as \textit{point correspondence constraints} in \cref{fig:interpolation}, is then plugged in the corresponding cost function $J^\mathit{corr}$:
\begin{equation}
    J^\mathit{corr} = \sum\nolimits_j \rho_i \bigl( e_{j, \mathit{corr}}^T \mathbf{\Omega}_\mathit{corr} e_{j, \mathit{corr}} \bigr) \, ,
\end{equation}
with $\mathbf{\Omega}_\mathit{corr}$ denoting the information matrix for all $j$.
As indicated in \cref{eqn:error-terms}, we do not add an error term for every image-point cloud pair. For instance, we do not apply $J^\mathit{corr}$ for all $p^\mathit{lidar}_j \in \mathit{Pcl}^\mathit{lidar}_j$ but only to a subset reducing the number of partially redundant constraints in the optimization problem.
Finally, we repeat the process of MAP estimation to obtain the overall calibration parameters $\mathbf{T}^\mathit{calib}_\mathit{final}$:
\begin{equation}
    \mathbf{T}^\mathit{calib}_\mathit{final} = \argmin_{\mathbf{T}^\mathit{calib}} J^\mathit{mot} + J^\mathit{corr} \, .
\end{equation}

%%%%%%%%%%%%%%%%%%%%%%%%%%%%%%%%%%%%%%%%%%%%%%%%%%%%%%%%%%%%%%%%%%%%%%%%%%%%%%%%

\subsection{Pose Synchronization}
\label{ssec:pose-interp}

\begin{figure}
    \centering
    \includegraphics[width=\linewidth]{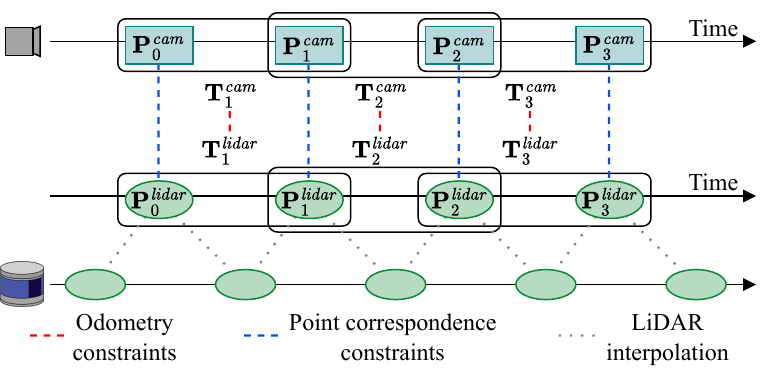}
    \vspace{-.6cm}
    \caption{We interpolate the poses from LiDAR odometry to the timestamps of the camera poses and further project the point cloud of the nearest neighbor to the same time to yield synchronized image-point cloud pairs. The odometry poses are then used to align the sensor motion, whereas the sensor measurements are fed to CMRNext~\cite{cattaneo2024cmrnext} to obtain point correspondences.}
    \label{fig:interpolation}
    \vspace{-.3cm}
\end{figure}

As noted in \cref{ssec:coarse-registration}, we interpolate camera and LiDAR poses to the same timestamp to yield synchronized sensor motion pairs. We further assume the usage of a global-shutter camera, i.e., all pixels are captured simultaneously. Since shifting a point cloud from one timestamp to another is significantly easier than simulating an image to be taken at a different time, we use the timestamps of the camera data as the reference. Next, we identify the two LiDAR measurements recorded before and after the image was taken, shown in \cref{fig:interpolation}. Based on the corresponding poses from LiDAR odometry, we compute an intermediate pose via linear and spherical linear interpolation for the translation and rotation components, respectively. Finally, we project the point cloud of the nearest neighbor to the same timestamp to obtain synchronized image-point cloud pairs that are used as input to CMRNext~\cite{cattaneo2024cmrnext}.

\section{Experimental Evaluation}

In this section, we extensively evaluate our proposed \net approach for various robotic platforms and provide a comparison with previous methods. We further analyze the effects of several parameters of our method.

%%%%%%%%%%%%%%%%%%%%%%%%%%%%%%%%%%%%%%%%%%%%%%%%%%%%%%%%%%%%%%%%%%%%%%%%%%%%%%%%

\subsection{Robotic Platforms}

We apply our method to a diverse set of sensor setups on four robotic platforms including the public KITTI dataset~\cite{geiger2013vision} as well as three in-house configurations. In the supplementary material, we provide further results on the Argoverse~2~\cite{wilson2021argoverse2} dataset.

\begin{table*}
\scriptsize
\centering
\caption{Calibration Error on the KITTI Dataset}
\vspace{-0.2cm}
\label{tab:results-kitti}
\setlength\tabcolsep{4.0pt}
\begin{threeparttable}
    \begin{tabular}{l c | cc !{\color{gray}\vline} ccc ccc | cc !{\color{gray}\vline} ccc ccc}
        \toprule
        & & \multicolumn{8}{c|}{\textbf{Left camera}} & \multicolumn{8}{c}{\textbf{Right camera}} \\
        & & \multicolumn{2}{c}{Magnitude} & \multicolumn{3}{c}{Translation [cm]} & \multicolumn{3}{c|}{Rotation [\degree]} & \multicolumn{2}{c}{Magnitude} & \multicolumn{3}{c}{Translation [cm]} & \multicolumn{3}{c}{Rotation [\degree]} \\
        \textbf{Method} & \textbf{Initial range} & $E_\textrm{t}$ [cm] & $E_\textrm{R}$ [\degree] & x & y & z & roll & pitch & yaw & $E_\textrm{t}$ [cm] & $E_\textrm{R}$ [\degree] & x & y & z & roll & pitch & yaw \\
        \midrule
        Tu~\textit{et~al.}\textsuperscript{\textdagger}~\cite{tu2022multicamera} & \SI{\pm0.20}{\meter} / \SI{\pm2}{\degree} & 4.40 & 0.16 & -- & -- & -- & -- & -- & -- & -- & -- & -- & -- & -- & -- & -- & -- \\[-1.25pt]
        Yin~\textit{et~al.}\textsuperscript{\textdagger}~\cite{yin2023automatic} & -- & 5.91 & 0.16 & 2.90 & 4.90 & 1.60 & 0.08 & 0.09 & 0.10 & -- & -- & -- & -- & -- & -- & -- & -- \\[-1.25pt]
        Borer~\textit{et~al.}\textsuperscript{\textdagger}~\cite{borer2024chaos} & \SI{\pm0.25}{\meter} / \SI{\pm1}{\degree} & 9.51 & 0.18 & 9.40 & 1.30 & 0.60 & 0.18 & \underline{0.03} & \textbf{0.03} & -- & -- & -- & -- & -- & -- & -- & -- \\[-1.25pt]
        CalibDepth\textsuperscript{\textdagger}~\cite{zhu2023calibdepth}&\SI{\pm1.5}{\meter} / \SI{\pm20}{\degree} & 1.17 & 0.12 & 1.31 & 1.02 & 1.17 & 0.06 & 0.23 & 0.08 & -- & -- & -- & -- & -- & -- & -- & -- \\
        CMRNet~\cite{cattaneo2019cmrnet} & \SI{\pm1.5}{\meter} / \SI{\pm20}{\degree} & 1.57 & 0.10 & 1.06 & 0.74 & \underline{0.34} & \underline{0.03} & \textbf{0.01} & 0.08 & 52.92 & 1.49 & \underline{1.59} & 52.87 & \textbf{0.36} & 0.04 & \textbf{0.02} & 1.49 \\
        RGGNet~\cite{yuan2020rggnet} & \SI{\pm0.3}{\meter} / \SI{\pm20}{\degree} & 11.49 & 1.29 & 8.14 & 2.79 & 3.97 & 0.35 & 0.74 & 0.64 & 23.52 & 3.87 & 18.03 & 5.55 & 6.06 & 0.51 & 3.38, & 1.48 \\
        LCCNet~\cite{lv2021lccnet} & \SI{\pm1.5}{\meter} / \SI{\pm20}{\degree} & \underline{1.01} & 0.12 & \underline{0.26} & \underline{0.36} & 0.35 & \textbf{0.02} & 0.11 & \textbf{0.03} & 52.51 & 1.47 & 52.48 & \textbf{0.26} & 0.74 & \textbf{0.01} & 1.47 & \textbf{0.03} \\
        CMRNext~\cite{cattaneo2024cmrnext} & \SI{\pm1.5}{\meter} / \SI{\pm20}{\degree} & 1.89 & \underline{0.08} & 1.12 & 0.83 & 0.79 & 0.04 & 0.04 & \underline{0.04} & \underline{7.07} & \underline{0.23} & 2.17 & 5.78 & 0.94 & 0.05 & \underline{0.05} & 0.20 \\
        \grayrule
        \net (\textit{ours}) & -- & \textbf{0.18} & \textbf{0.06} & \textbf{0.07} & \textbf{0.16} & \textbf{0.01} & \textbf{0.02} & 0.04 & \underline{0.04} & \textbf{2.94} & \textbf{0.14} & \textbf{0.66} & \underline{2.78} & \underline{0.49} & \underline{0.03} & \underline{0.05} & \underline{0.13} \\
        \bottomrule
    \end{tabular}
    \footnotesize
    We provide results based on data from sequence 00 of the KITTI odometry benchmark~\cite{geiger2013vision}. Unlike many other works, we evaluate our approach and previous methods for both cameras.
    Bold and underlined values indicate the best and second-best scores, respectively.
    \textdagger: These methods did not release (English-speaking) code preventing reproducing results for both cameras.
\end{threeparttable}
\vspace*{-0.4cm}
\end{table*}

\begin{figure}
    \centering
    \includegraphics[width=\linewidth,trim={7cm 0 0 0},clip]{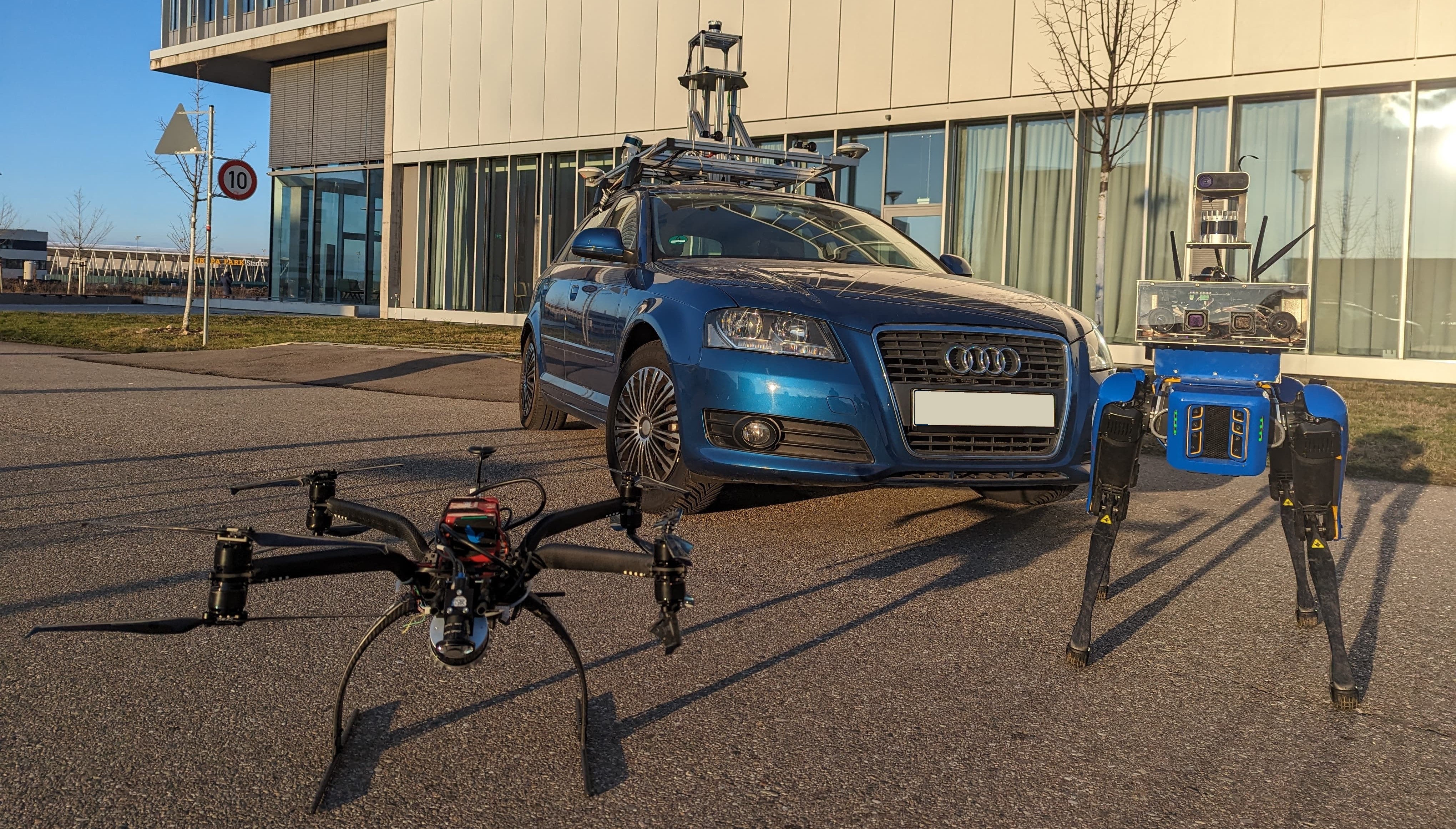}
    \vspace{-.6cm}
    \caption{We calibrate the sensors on three in-house robotic platforms including a self-driving perception car, a quadruped robot, and a UAV.}
    \label{fig:vehicle-spot}
    \vspace{-.4cm}
\end{figure}

{\parskip=3pt
\noindent\textit{KITTI Dataset:}
We extensively evaluate our method on the publicly available KITTI dataset~\cite{geiger2013vision}. In detail, we utilize the rectified images and undistorted point clouds of sequence~00 of the odometry benchmark as data from the left camera of the other sequences was seen during the training of CMRNext~\cite{cattaneo2024cmrnext}. The vehicle is equipped with two RGB cameras Point Grey Flea~2 and one LiDAR of type Velodyne HDL-64E. Sensor data is captured at \SI{10}{\hertz}.
}

{\parskip=3pt
\noindent\textit{In-House Robots:}
We further employ our method to three in-house datasets that were not seen during the training of CMRNext~\cite{cattaneo2024cmrnext} and include new sensor models. Each robot, shown in \cref{fig:vehicle-spot}, is equipped with one Ouster \mbox{OS-1} LiDAR with 128 channels and different cameras. In detail, our vehicle is equipped with four FLIR Blackfly 2353C, the quadruped robot includes five AVT cameras, and the UAV uses a FLIR Blackfly S. The sensors on all robots capture data at \SI{10}{\hertz}. The LiDAR is time synchronized via PTP, whereas the images are stamped on arrival. To compute the calibration error, we obtain reference parameters using manually selected camera-LiDAR point correspondences~\cite{gtcalibration}.
}

%%%%%%%%%%%%%%%%%%%%%%%%%%%%%%%%%%%%%%%%%%%%%%%%%%%%%%%%%%%%%%%%%%%%%%%%%%%%%%%%

\subsection{Main Results}

We measure the calibration error as the difference between the calibration parameters produced by \net and the provided or manually obtained parameters. Throughout this section, we report both the magnitude of the rotation and translation errors as well as the errors of the individual axes. In detail, we compute the errors as follows:
\begin{subequations}
\begin{align}
    E_\textrm{t} &= || \mathbf{t} - \mathbf{\hat t} ||_2 \, , \\
    m            &= q \ast \hat q^{-1} \, , \\
    E_\textrm{R} &= \atantwo \left( \sqrt{m_x^2 + m_y^2 + m_z^2}, m_w \right) \, ,
\end{align}
\end{subequations}
where $\mathbf{t}$ and $\mathbf{\hat t}$ are ground truth and predicted translation parameters, $q$ and $\hat q$ denote the corresponding parameters of the rotation $\mathbf{R}$ as quaternions. Finally, $\ast$ and ${}^{-1}$ are the multiplicative and inverse operations for quaternions.

For KITTI, we compare our \net approach to several classical~\cite{tu2022multicamera, yin2023automatic} and learning-based~\cite{zhu2023calibdepth, borer2024chaos, cattaneo2024cmrnext, lv2021lccnet, cattaneo2019cmrnet, yuan2020rggnet} baselines in \cref{tab:results-kitti}. If the authors released the corresponding code, we reproduce the calibration results for both the left and the right camera. Besides our \net, only the method by Yin~\textit{et~al.}~\cite{yin2023automatic} is initialization-free. For the others, we follow the original sampling ranges. Note that the reproduced metrics of the learning-based approaches correspond to median calibration over multiple frames.
For our method, we utilize 1,000 sensor motion constraints and 5\% randomly sampled correspondences from each of the 100 image-point cloud pairs. 
Although all baselines are outperformed by our approach, they generally yield accurate results for calibrating the LiDAR to the left camera. However, it is paramount to emphasize that all learning-based methods incorporate samples of the left camera in their training data. Therefore, we also calibrate the right camera measuring the capability to generalize. Except for CMRNext, all baselines suffer from a substantial performance drop. Nonetheless, \net yields the smallest error demonstrating that our joint optimization further increases robustness to unseen sensor configuration.

We confirm this observation by employing \net, CMRNext~\cite{cattaneo2024cmrnext}, and the method by Koide~\textit{et~al.}~\cite{koide2023general} to three in-house robotic platforms. To account for the missing hardware time synchronization between camera and LiDAR, we only utilize static frames for computing point correspondences. As shown in \cref{tab:results-inhouse}, \net yields significantly smaller calibration errors on the vehicle and the quadruped robot. For the UAV, we hypothesize that point cloud undistortion is more challenging due to less constrained motion resulting in less accurate odometry and hence larger errors.

\begin{figure*}[t]
    \centering
    \includegraphics[width=\linewidth]{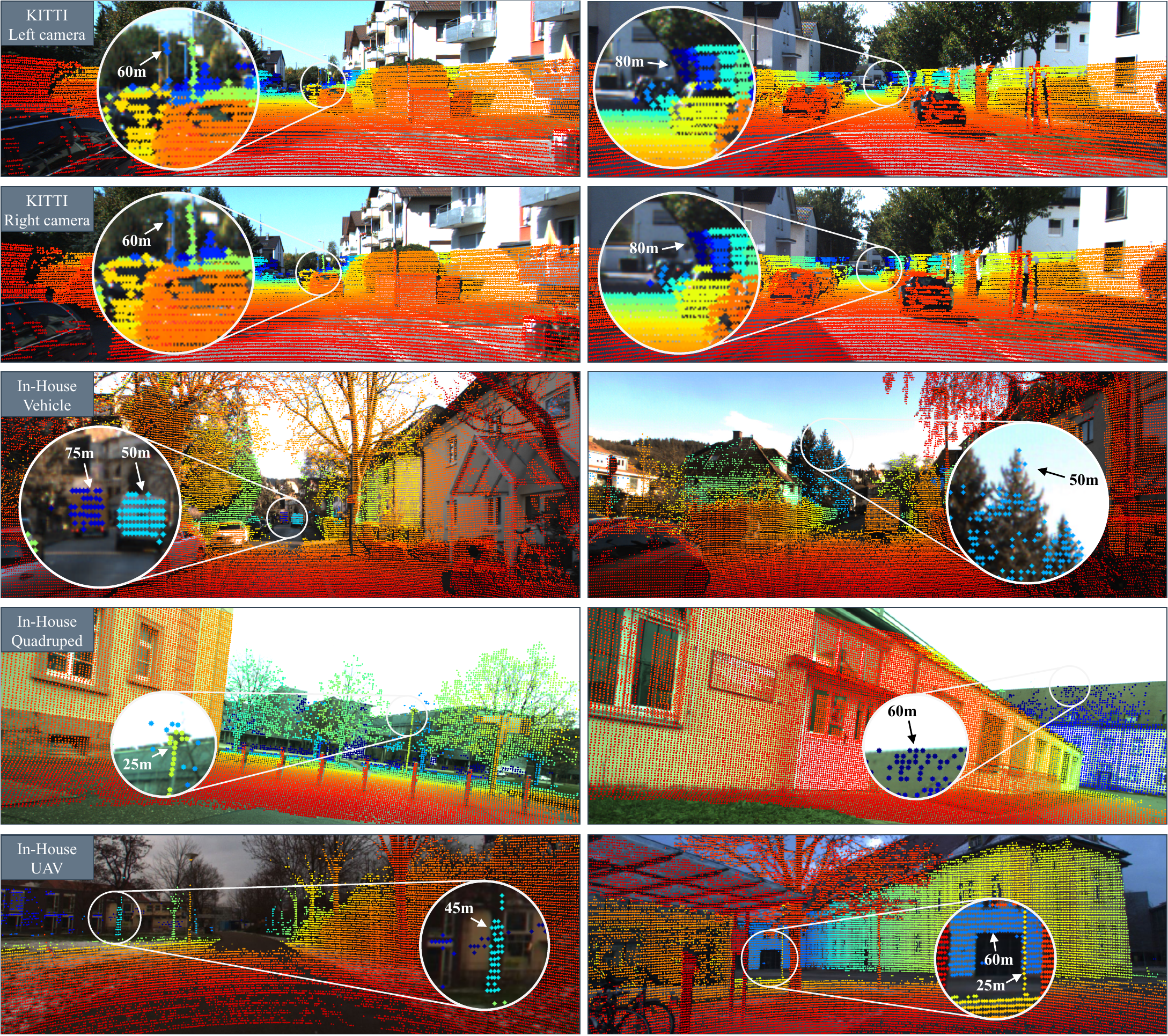}
    \vspace*{-.7cm}
    \caption{Qualitative results of \net on the KITTI dataset and our in-house robotic platforms. Best viewed on screen and zoomed in.}
    \label{fig:results}
    \vspace*{-.6cm}
\end{figure*}

We report our calibration parameters for KITTI in the supplementary material. We show qualitative results by projecting the point clouds onto the images in \cref{fig:results} and provide further visualizations in the complementary video. Notably, the projections based on our calibration are still highly accurate at the maximum distance of \SI{80}{\meter}.

\begin{table}
\scriptsize
\centering
\caption{Calibration Error on In-House Robotic Platforms}
\vspace{-0.2cm}
\label{tab:results-inhouse}
\setlength\tabcolsep{1.8pt}
\begin{threeparttable}
    \begin{tabular}{cl | cc !{\color{gray}\vline} ccc ccc}
        \toprule
        & & \multicolumn{2}{c}{Magnitude} & \multicolumn{3}{c}{Translation [cm]} & \multicolumn{3}{c}{Rotation [\degree]} \\
        \textbf{Platform} & \textbf{Method} & $E_\textrm{t}$ [cm] & $E_\textrm{R}$ [\degree] & x & y & z & roll & pitch & yaw \\
        \midrule
        \multirow{3}{*}{Vehicle} & Koide~\textit{et~al.}~\cite{koide2023general} & \underline{9.76}  & \textbf{0.27} & 9.24 & \textbf{0.63} & \underline{3.06} & \underline{0.07} & \textbf{0.18} & \underline{0.19}  \\
        & CMRNext~\cite{cattaneo2024cmrnext} & 12.20 & \underline{0.90} & \textbf{1.35} & 10.83 & 3.60 & 0.45 & 0.82 & 0.38 \\
        & \net (\textit{ours}) & \textbf{4.50} & \textbf{0.27} & \underline{4.04} & \underline{1.71} & \textbf{1.03} & \textbf{0.00} & \underline{0.19} & \textbf{0.18} \\  
        \midrule
        \multirow{3}{*}{Quadruped} & Koide~\textit{et~al.}~\cite{koide2023general} & \underline{16.21} &  \underline{1.34} & \underline{2.10} & \underline{15.61} & \textbf{3.84} &  1.07 & \underline{0.67} & 0.44 \\
        & CMRNext~\cite{cattaneo2024cmrnext} & 23.95 & 1.36 & 3.73 & 16.07 & 15.05 & \underline{0.40} & 0.85 & \underline{0.32} \\
        & \net (\textit{ours}) & \textbf{9.54} & \textbf{0.38} & \textbf{1.27} & \textbf{2.70} & \underline{9.06} & \textbf{0.31} & \textbf{0.19} & \textbf{0.10} \\  
        \midrule
        \multirow{3}{*}{UAV} & Koide~\textit{et~al.}~\cite{koide2023general} &  \textbf{1.65} &  \textbf{0.36} & \underline{1.56} &  \textbf{0.08} &  \textbf{0.50} & \underline{0.16} & \underline{0.17} &  \textbf{0.28}  \\
        & CMRNext~\cite{cattaneo2024cmrnext} & 12.47 & 0.97 &  \textbf{1.17} & 5.31 & 6.83 & 0.27 & 0.34 & 0.36 \\
        & \net (\textit{ours}) & \underline{5.12} & \underline{0.51} & 4.19 & \underline{0.59} & \underline{2.88} &  \textbf{0.15} &  \textbf{0.13} & \underline{0.30} \\
        \bottomrule
    \end{tabular}
    \footnotesize
    We obtain reference calibration parameters to compute the errors by using manually selected camera-LiDAR point correspondences~\cite{gtcalibration}. Neither our \net nor the method by Koide~\textit{et~al.}~\cite{koide2023general} requires an initial guess.
    Bold and underlined values indicate the best and second-best scores.
\end{threeparttable}
\vspace{-0.4cm}
\end{table}

%%%%%%%%%%%%%%%%%%%%%%%%%%%%%%%%%%%%%%%%%%%%%%%%%%%%%%%%%%%%%%%%%%%%%%%%%%%%%%%%

\subsection{Ablation Studies}
\label{ssec:ablation-studies}

\begin{figure}[t]
    \centering
    \includegraphics[width=\linewidth]{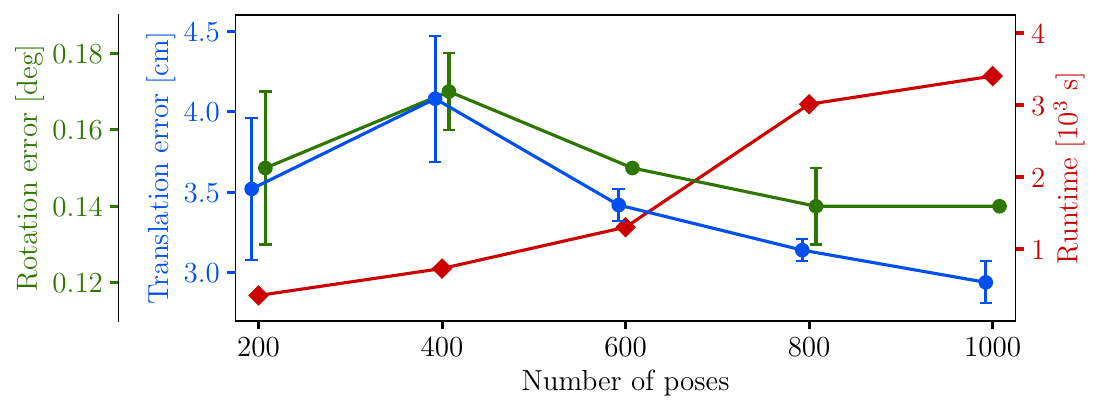}
    \vspace{-.8cm}
    \caption{For the point correspondence constraints, we use 5\% correspondences per image-point cloud pair from a total of 100 pairs.
    }
    \label{fig:ablation-poses}
    \vspace{-.3cm}
\end{figure}

\begin{figure}[t]
    \centering
    \includegraphics[width=\linewidth]{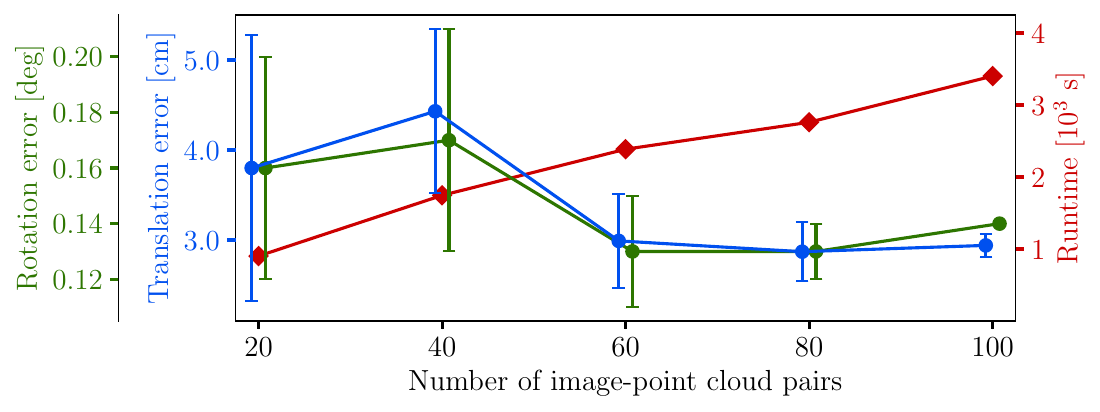}
    \vspace{-.8cm}
    \caption{For the motion constraints, we use 1,000 matched poses. Per image-point cloud pair, we select 5\% of the correspondences.
    }
    \label{fig:ablation-correspondences}
    \vspace{-.5cm}
\end{figure}

\begin{figure}[t]
    \centering
    \includegraphics[width=\linewidth]{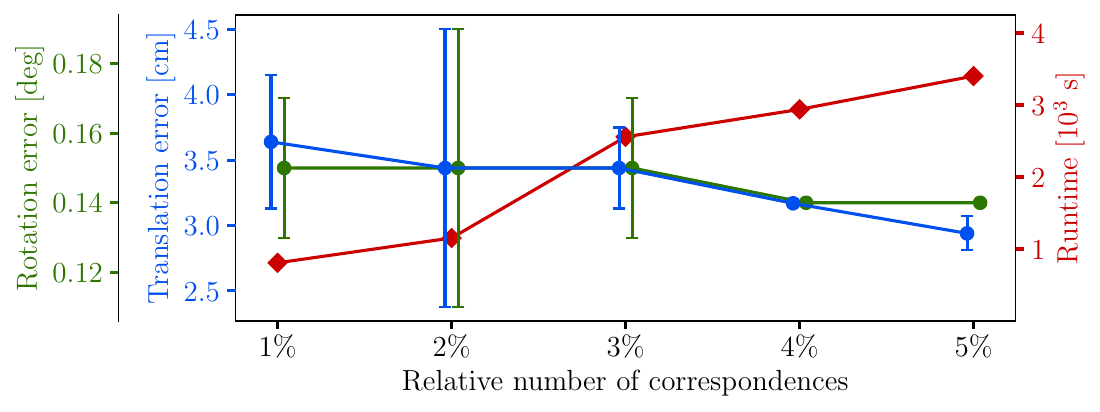}
    \vspace{-.9cm}
    \caption{For the motion constraints, we use 1,000 matched poses. We extract correspondences from a total of 100 image-point cloud pairs.
    }
    \label{fig:ablation-pairs}
    \vspace{-.3cm}
\end{figure}

\begin{figure}[t]
    \centering
    \includegraphics[width=\linewidth]{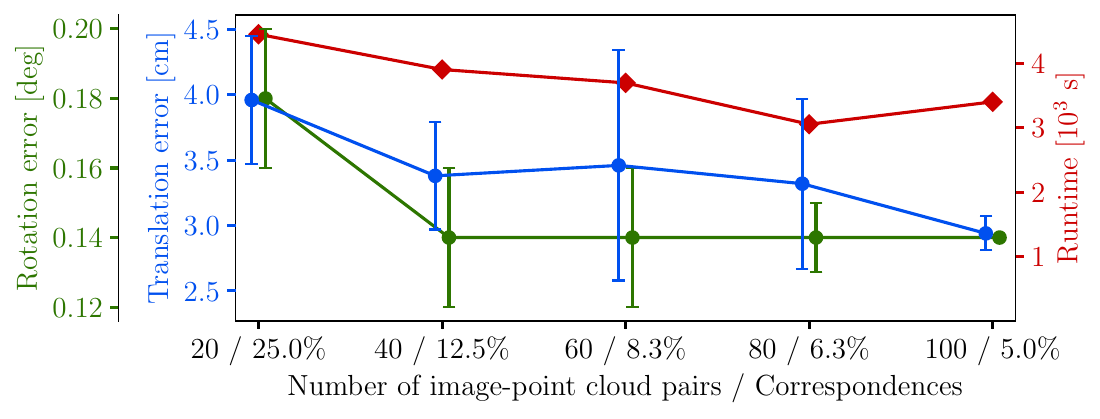}
    \vspace{-.9cm}
    \caption{For the motion constraints, we use 1,000 matched poses. We vary the pose count and the relative number of correspondences to obtain an approximately constant absolute number of correspondences.
    }
    \label{fig:ablation-pairs-correspondences}
    \vspace{-.5cm}
\end{figure}

We extensively analyze the sensitivity of \net towards several design choices and adjustable parameters by conducting various ablation studies on the KITTI dataset~\cite{geiger2013vision}. In particular, we present results for calibrating the right RGB camera based on sequence 00. If not explicitly stated otherwise, we utilize 1,000 odometry poses and 5\% of the point correspondences from 100 image-point cloud pairs. In \cref{fig:ablation-poses,fig:ablation-pairs,fig:ablation-correspondences,fig:ablation-pairs-correspondences}, we visualize the mean and standard deviation of three runs. To improve readability, we add small offsets to the $x$-values. For convenience, we provide the raw numerical metrics in the supplementary material.

{\parskip=3pt
\noindent\textit{Components Analysis:}
We analyze the impact of the calibration stages and the components of the optimization problem in \cref{tab:components-analysis}.
Although the initialization-free coarse registration step based on sensor motion reduces the rotation error to sub-degree accuracy, the translation parameters suffer from a lack of observability.
For reference, we provide the error of the median calibration parameters when running PnP on 100\% of the predicted point correspondences. In \cref{tab:components-analysis}, we refer to this as CMRNext~\cite{cattaneo2024cmrnext}.
In \net, we instead process a subset of the correspondences as constraints in the optimization formulation, which further reduces the errors. 
In the bottom two rows, we demonstrate the efficacy of the key ingredients of \net. First, we show that utilizing the Cauchy loss~\cite{liu2014cauchy} to robustify the cost functions significantly improves the results by reducing the effect of outliers in the observations. Second, while previous works~\cite{taylor2015motion, yin2023automatic} only proposed to utilize sensor motion to initialize a subsequent correspondence-based calibration scheme, we incorporate both constraints in the refinement stage leveraging their complementary information. Our experiment clearly underlines the positive impact of joint optimization with respect to both sensor motion and point correspondences.
}

{\parskip=3pt
\noindent\textit{Runtime Analysis:}
We discuss the runtime of our method with respect to the number of both motion-based and point correspondences-based constraints. We conduct these experiments on a machine with an AMD Ryzen Threadripper PRO 3975WX CPU with \SI{128}{\giga\byte} and an NVIDIA A6000 GPU with \SI{48}{\giga\byte}. In \cref{fig:ablation-poses} and \cref{fig:ablation-correspondences}, we visualize the runtime versus the number of poses and the relative number of correspondences, respectively. As can be seen in both studies, the runtime scales approximately linearly, whereas the errors decrease. We further observe an optimal configuration, after which the runtime continues to increase without major impacts on the accuracy.
}

\begin{table}
\scriptsize
\centering
\caption{Components Analysis}
\vspace{-0.2cm}
\label{tab:components-analysis}
\setlength\tabcolsep{3.0pt}
\begin{threeparttable}
    \begin{tabular}{l | cc !{\color{gray}\vline} ccc ccc}
        \toprule
        & \multicolumn{2}{c}{Magnitude} & \multicolumn{3}{c}{Translation [cm]} & \multicolumn{3}{c}{Rotation [\degree]} \\
        \textbf{Component} & $E_\textrm{t}$ [cm] & $E_\textrm{R}$ [\degree] & x & y & z & roll & pitch & yaw \\
        \midrule
        Coarse registration & 39.37 & 0.51 & 18.11 & 3.57 & 34.77 & 0.38 & 0.33 & \textbf{0.09} \\
        \hspace{.5pt} + CMRNext~\cite{cattaneo2024cmrnext} & 6.26 & 0.28 & \textbf{0.31} & 6.25 & \textbf{0.00} & \textbf{0.01} & \textbf{0.00} & 0.28\\
        Fine registration: & & & \\
        \hspace{.5pt} Point constraints  & 5.89 & 0.21 & 0.78 & 5.81 & 0.50 & 0.04 & 0.06 & 0.19 \\
        \hspace{.5pt} + Cauchy robustifier & \underline{3.42} & \underline{0.16} & \underline{0.35} & \underline{3.38} & \underline{0.37} & \underline{0.02} & \underline{0.05} & 0.15 \\
        \rowcolor{Gray}
        \hspace{.5pt} + Motion constraints & \textbf{2.94} & \textbf{0.14} & 0.66 & \textbf{2.78} & 0.49 & 0.03 & \underline{0.05} & \underline{0.13} \\
        \bottomrule
    \end{tabular}
    \footnotesize
    The last line highlighted in gray corresponds to our proposed \net.
    CMRNext denotes the errors of the median calibration parameters when running PnP on 100\% of the predicted point correspondences after initialization with the coarse registration.
    Bold and underlined values indicate the best and second-best scores, respectively.
\end{threeparttable}
\vspace{-0.9cm}
\end{table}

{\parskip=3pt
\noindent\textit{Number of Poses:}
In this study, we evaluate the effect of the number of poses from visual and LiDAR odometry on the calibration error, i.e., the length of the required data recording. In \cref{fig:ablation-poses}, we show the translation and rotation errors for an increasing number of poses. Although it generally holds that adding more motion constraints to the optimization problem results in more accurate calibration parameters, we observe that the error converges.
}

{\parskip=3pt
\noindent\textit{Number of Point Correspondences:}
Next, we repeat a similar study for the number of constraints induced by the point correspondences. For KITTI data, we measure an average of 20,000 correspondences per image-point cloud pair, i.e., our default setting of utilizing 5\% generates approximately 1,000 constraints. In \cref{fig:ablation-correspondences}, we keep a fixed number of 100 pairs and vary the relative number of constraints. Fewer correspondences result in higher calibration errors. We confirm this observation in a second experiment visualized in \cref{fig:ablation-pairs}, where we utilize 5\% of the point correspondences but reduce the number of image-point cloud pairs. Note that the errors converge towards a lower limit in both experimental setups. Finally, we address the following question: Given a fixed number of correspondences, and hence runtime, is it preferable to increase the number of image-point cloud pairs or the relative amount of correspondences per pair? In \cref{fig:ablation-pairs-correspondences}, we plot the calibration errors for a stable number of point correspondences, i.e., when decreasing the number of pairs with respect to the default setting, we increase the relative number of correspondences by the respective amount. The experiment shows that a smaller number of correspondences from a more diverse set of data pairs is preferable.
}

\section{Conclusion}

In this paper, we proposed \net, a novel method for the extrinsic calibration between camera and LiDAR sensors in an automated manner without the need for dedicated calibration targets. Our approach utilizes non-linear optimization to obtain the calibration parameters by aligning sensor motions from visual and LiDAR odometry and leveraging deep learning-based correspondences between 2D pixels and 3D points.
In contrast to most previous learning-based methods, our approach generalizes to sensor configurations that differ from the training setup. Importantly, \net does not require accurate parameters for initialization.
In practical experiments carried out on diverse robotic platforms, we demonstrated the efficacy of our method and provided detailed evaluations of several design choices.
Future versions of our approach could incorporate constraints for multiple cameras and LiDARs and include intrinsic calibration in the optimization problem. Due to its unsupervised procedure, MDPCalib could further be extended to online calibration.

\section*{Acknowledgment}

The authors thank Johan Vertens for designing the in-house driving platform used in this work and both Julia Hindel and Iana Zhura for assistance with the quadrupedal robot. Finally, the authors thank Wera Winterhalter, Freya Fleckenstein, and Srdjan Milojevic for technical discussions.

%%%%%%%%%%%%%%%%%%%%%%%%%%%%%%%%%%%%%%%%%%%%%%%%%%%%%%%%%%%%%%%%%%%%%%%%%%%%%%%%

\footnotesize
\bibliographystyle{IEEEtran}
\bibliography{references.bib}

%%%%%%%%%%%%%%%%%%%%%%%%%%%%%%%%%%%%%%%%%%%%%%%%%%%%%%%%%%%%%%%%%%%%%%%%%%%%%%%%

%%%%%%%%%% Merge with supplemental materials %%%%%%%%%%
\clearpage
\renewcommand{\baselinestretch}{1}
\setlength{\belowcaptionskip}{0pt}

\sisetup{output-exponent-marker=\ensuremath{\mathrm{e}}}

\begin{strip}
\begin{center}
\vspace{-5ex}

\textbf{\LARGE \bf
Automatic Target-Less Camera-LiDAR Calibration \\\vspace{.5ex} From Motion and Deep Point Correspondences} \\
\vspace{3ex}

\Large{\bf- Supplementary Material -}\\
 \vspace{0.4cm}
 \normalsize{
Kürsat Petek$^{1*}$,
Niclas Vödisch$^{1*}$,
Johannes Meyer$^{1}$,
Daniele Cattaneo$^{1}$,
Abhinav Valada$^{1}$,
Wolfram Burgard$^{2}$}
\end{center}
\end{strip}

%%%%%%%%%% Merge with supplemental materials %%%%%%%%%%
\setcounter{section}{0}
\setcounter{equation}{0}
\setcounter{figure}{0}
\setcounter{table}{0}
\setcounter{page}{1}
\makeatletter

%%%%%%%%%% Prefix a "S" to all equations, figures, tables and reset the counter %%%%%%%%%%

\renewcommand{\thesection}{S-\Roman{section}}
\renewcommand{\thesubsection}{S-\arabic{subsection}}
\renewcommand{\thetable}{S-\Roman{table}}
\renewcommand{\thefigure}{S-\arabic{figure}}
\renewcommand{\theequation}{S-\arabic{equation}}

 \let\thefootnote\relax\footnote{$^{*}$ Equal contribution.\\
 $^{1}$ Department of Computer Science, University of Freiburg, Germany.\\
 $^{2}$ Department of Eng., University of Technology Nuremberg, Germany.
 }%
\normalsize

%%%%%%%%%%%%%%%%%%%%%%%%%%%%%%%%%%%%%%%%%%%%%%%%%%%%%%%%%%%%%%%%%%%%%%%%%%%%%%%%

In this supplementary material, we report results on the Argoverse~2~\cite{wilson2021argoverse2} dataset, release calibration parameters for the KITTI dataset~\cite{geiger2013vision}, and provide more detailed numbers for our ablation studies.

%%%%%%%%%%%%%%%%%%%%%%%%%%%%%%%%%%%%%%%%%%%%%%%%%%%%%%%%%%%%%%%%%%%%%%%%%%%%%%%%

\section{Results on Argoverse~2}

In this section, we provide results on the Argoverse~2~\cite{wilson2021argoverse2} dataset. In contrast to the KITTI odometry benchmark, Argoverse~2 contains many short sequences lasting only 15-30s, which is too short for our method to be applied. Furthermore, since the ground truth extrinsic calibration is not consistent between all sequences, we collect sequences that share the same set of ground truth parameters and combine them in a larger sequence with small gaps in between that can be detected by our method via the lost-track feature of ORB-SLAM3~\cite{campos2021orbslam}. In particular, in this experiment, we utilize sequence \texttt{05fb81ab-5e46-3f63-a59f-82fc66d5a477} as the reference sequence. Note that CMRNext~\cite{cattaneo2024cmrnext} has not been trained on the Argoverse~2~\cite{wilson2021argoverse2} dataset, therefore, allowing us to use sequences from the \texttt{train} split for evaluation. In \cref{tab:results-argoverse}, we report the calibration error for both stereo cameras of CMRNext~\cite{cattaneo2024cmrnext} and our method \net.

%%%%%%%%%%%%%%%%%%%%%%%%%%%%%%%%%%%%%%%%%%%%%%%%%%%%%%%%%%%%%%%%%%%%%%%%%%%%%%%%

\section{KITTI Calibration Parameters}

To foster future research towards camera-LiDAR sensor fusion, we release the calibration parameters that we obtained with our method for sequence 00 of the KITTI dataset~\cite{geiger2013vision}.
In detail, we provide the rotation matrices and translation vectors for projecting the LiDAR point cloud into the image.

{\parskip=3pt
\noindent\textit{Left RGB Camera:}
% \begin{subequations}
\begin{align*}
    \mathbf{R} & = 
    \begin{bmatrix*}[r]
    \num{-1.2619e-04} & \num{-9.9997e-01} & \num{-8.2230e-03} \\
    \num{-7.8537e-03} & \num{8.2238e-03} & \num{-9.9994e-01} \\
    \num{9.9997e-01} & \num{-6.1602e-05} & \num{-7.8545e-03}
    \end{bmatrix*} \\
    \mathbf{t} & = 
    \begin{bmatrix*}[r]
    \num{5.1090e-02} & \num{-5.5873e-02} & \num{-2.9575e-01}
    \end{bmatrix*}^T
\end{align*}
% \end{subequations}
}

{\parskip=3pt
\noindent\textit{Right RGB Camera:}
% \begin{subequations}
\begin{align*}
    \mathbf{R} & = 
    \begin{bmatrix*}[r]
    \num{-1.8202e-03} & \num{-9.9996e-01} & \num{-8.2416e-03} \\
    \num{-8.2823e-03} & \num{8.2564e-03} & \num{-9.9993e-01} \\
    \num{9.9996e-01} & \num{-1.7518e-03} & \num{-8.2970e-03} \\
    \end{bmatrix*} \\
    \mathbf{t} & = 
    \begin{bmatrix*}
    \num{-4.5166e-01} & \num{-4.8448e-02} & \num{-2.8787e-01}
    \end{bmatrix*}^T
\end{align*}
% \end{subequations}
}

\begin{table}[t]
% \footnotesize
\scriptsize
\centering
\caption{Calibration Error on the Argoverse~2 Dataset}
\vspace{-0.2cm}
\label{tab:results-argoverse}
\setlength\tabcolsep{3.5pt}
\begin{threeparttable}
    \begin{tabular}{l | cc !{\color{gray}\vline} ccc ccc}
        \toprule
        & \multicolumn{8}{c}{\textbf{Stereo left camera}} \\
        & \multicolumn{2}{c}{Magnitude} & \multicolumn{3}{c}{Translation [cm]} & \multicolumn{3}{c}{Rotation [\degree]} \\
        \textbf{Method} & $E_\textrm{t}$ [cm] & $E_\textrm{R}$ [\degree] & x & y & z & roll & pitch & yaw \\
        \midrule
        CMRNext~\cite{cattaneo2024cmrnext} & 17.31 & 0.35 & 10.52 & 10.82 & 8.47 & \textbf{0.01} & 0.30 & 0.18 \\
        MDPCalib (\textit{ours}) & \textbf{9.48} & \textbf{0.18} & \textbf{7.86} & \textbf{5.03} & \textbf{1.70} & 0.04 & \textbf{0.15} & \textbf{0.09} \\
        \bottomrule
        \multicolumn{2}{c}{} \\[.1cm]
        \toprule
        & \multicolumn{8}{c}{\textbf{Stereo right camera}} \\
        & \multicolumn{2}{c}{Magnitude} & \multicolumn{3}{c}{Translation [cm]} & \multicolumn{3}{c}{Rotation [\degree]} \\
        \textbf{Method} & $E_\textrm{t}$ [cm] & $E_\textrm{R}$ [\degree] & x & y & z & roll & pitch & yaw \\
        \midrule
        CMRNext~\cite{cattaneo2024cmrnext} & 31.27 & 0.36 & 29.15 & 9.92 & 5.44 & 0.14 & 0.29 & 0.15 \\
        MDPCalib (\textit{ours}) & \textbf{22.90} & \textbf{0.21} & \textbf{20.88} & \textbf{8.86} & \textbf{3.17} & \textbf{0.01} & \textbf{0.17} & \textbf{0.12} \\
        \bottomrule
    \end{tabular}
    \footnotesize
    We use sequence \texttt{05fb81ab-5e46-3f63-a59f-82fc66d5a477} and sequences with the same set of ground truth calibration parameters. Bold values denote the best score per metric.
\end{threeparttable}
% \vspace*{-0.2cm}
\end{table}

%%%%%%%%%%%%%%%%%%%%%%%%%%%%%%%%%%%%%%%%%%%%%%%%%%%%%%%%%%%%%%%%%%%%%%%%%%%%%%%%

\section{Ablation Studies}

In this section, we provide the numerical values that are used to generate the figures in the main manuscript.
Throughout the tables, we highlight the parameters that correspond to our overall setting in gray. For the magnitude errors $E_\textrm{t}$ and $E_\textrm{R}$, we report the mean and standard deviation of three runs. For the errors of the individual axes, we report only the mean.

\vfill\null
\newpage

\begin{table}
% \footnotesize
\scriptsize
\centering
\caption{Number of Poses}
\vspace{-0.2cm}
\label{tab:ablation-poses}
\setlength\tabcolsep{3.0pt}
\begin{threeparttable}
    \begin{tabular}{c | cc !{\color{gray}\vline} ccc ccc | c}
        \toprule
        \textbf{Pose} & \multicolumn{2}{c}{Magnitude} & \multicolumn{3}{c}{Translation [cm]} & \multicolumn{3}{c|}{Rotation [\degree]} & Time [s] \\
        \textbf{count} & $E_\textrm{t}$ [cm] & $E_\textrm{R}$ [\degree] & x & y & z & roll & pitch & yaw \\
        \midrule
        \rowcolor{Gray}
        1000 & 2.94$\pm$0.13 & 0.14$\pm$0.00 & 0.66 & 2.78 & 0.49 & 0.03 & 0.05 & 0.13 & 3402 \\
         800 & 3.14$\pm$0.07 & 0.14$\pm$0.01 & 0.53 & 3.04 & 0.42 & 0.02 & 0.05 & 0.13 & 3011 \\
         600 & 3.42$\pm$0.10 & 0.15$\pm$0.00 & 0.08 & 3.39 & 0.36 & 0.02 & 0.05 & 0.14 & 1303 \\
         400 & 4.08$\pm$0.39 & 0.17$\pm$0.01 & 0.17 & 4.06 & 0.28 & 0.01 & 0.04 & 0.16 & 729\\
         200 & 3.52$\pm$0.44 & 0.15$\pm$0.02 & 0.68 & 3.44 & 0.23 & 0.02 & 0.04 & 0.14 & 355\\
        \bottomrule
    \end{tabular}
    \footnotesize
    This table corresponds to \cref{fig:ablation-poses}.
    For the point correspondence constraints, we use 5\% correspondences per image-point cloud pair from 100 pairs.
\end{threeparttable}
% \vspace{-0.5cm}
\end{table}

\begin{table}
% \footnotesize
\scriptsize
\centering
\caption{Relative Number of Point Correspondences}
\vspace{-0.2cm}
\label{tab:ablation-correspondences}
\setlength\tabcolsep{3.0pt}
\begin{threeparttable}
    \begin{tabular}{c | cc !{\color{gray}\vline} ccc ccc | c}
        \toprule
        \textbf{Corr.} & \multicolumn{2}{c}{Magnitude} & \multicolumn{3}{c}{Translation [cm]} & \multicolumn{3}{c|}{Rotation [\degree]} & Time [s] \\
        \textbf{count} & $E_\textrm{t}$ [cm] & $E_\textrm{R}$ [\degree] & x & y & z & roll & pitch & yaw \\
        \midrule
        \rowcolor{Gray}
        5\% & 2.94$\pm$0.13 & 0.14$\pm$0.00 & 0.66 & 2.78 & 0.49 & 0.03 & 0.05 & 0.13 & 3402 \\
        4\% & 3.17$\pm$0.03 & 0.14$\pm$0.00 & 0.39 & 3.12 & 0.34 & 0.01 & 0.04 & 0.13 & 2939 \\
        3\% & 3.44$\pm$0.31 & 0.15$\pm$0.02 & 0.50 & 3.36 & 0.43 & 0.02 & 0.05 & 0.14 & 2558 \\
        2\% & 3.44$\pm$1.06 & 0.15$\pm$0.04 & 0.19 & 3.42 & 0.34 & 0.01 & 0.04 & 0.08 & 1153 \\
        1\% & 3.64$\pm$0.51 & 0.15$\pm$0.02 & 0.18 & 3.59 & 0.31 & 0.02 & 0.05 & 0.14 & 808 \\
        \bottomrule
    \end{tabular}
    \footnotesize
    This table corresponds to \cref{fig:ablation-correspondences}.
    For the motion constraints, we use 1,000 poses. We extract correspondences from 100 image-point cloud pairs.
\end{threeparttable}
% \vspace{-0.5cm}
\end{table}

\vfill\null
\newpage

\begin{table}
% \footnotesize
\scriptsize
\centering
\caption{Number of Image-Point Cloud Pairs}
\vspace{-0.2cm}
\label{tab:ablation-pairs}
\setlength\tabcolsep{3.0pt}
\begin{threeparttable}
    \begin{tabular}{c | cc !{\color{gray}\vline} ccc ccc | c}
        \toprule
        \textbf{Pair} & \multicolumn{2}{c}{Magnitude} & \multicolumn{3}{c}{Translation [cm]} & \multicolumn{3}{c|}{Rotation [\degree]} & Time [s] \\
        \textbf{count} & $E_\textrm{t}$ [cm] & $E_\textrm{R}$ [\degree] & x & y & z & roll & pitch & yaw\\
        \midrule
        \rowcolor{Gray}
        100 & 2.94$\pm$0.13 & 0.14$\pm$0.00 & 0.66 & 2.78 & 0.49 & 0.03 & 0.05 & 0.13 & 3402 \\
         80 & 2.87$\pm$0.33 & 0.13$\pm$0.01 & 0.43 & 2.78 & 0.41 & 0.02 & 0.05 & 0.12 & 2760 \\
         60 & 2.99$\pm$0.52 & 0.13$\pm$0.02 & 0.10 & 2.96 & 0.32 & 0.02 & 0.05 & 0.12 & 2386 \\
         40 & 4.43$\pm$0.91 & 0.17$\pm$0.04 & 0.60 & 4.35 & 0.49 & 0.01 & 0.05 & 0.16 & 1745 \\
         20 & 3.80$\pm$1.48 & 0.16$\pm$0.04 & 0.69 & 3.60 & 0.12 & 0.01 & 0.05 & 0.15 & 902 \\
        \bottomrule
    \end{tabular}
    \footnotesize
    This table corresponds to \cref{fig:ablation-pairs}.
    For the motion constraints, we use 1,000 poses. Per image-point cloud pair, we select 5\% of the correspondences.
\end{threeparttable}
% \vspace{-0.5cm}
\end{table}

\begin{table}
% \footnotesize
\scriptsize
\centering
\caption{Pair Diversity vs. Relative Correspondences}
\vspace{-0.2cm}
\label{tab:ablation-pairs-correspondences}
\setlength\tabcolsep{2.2pt}
\begin{threeparttable}
    \begin{tabular}{c | cc !{\color{gray}\vline} ccc ccc | c}
        \toprule
        \textbf{Pair / Corr.} & \multicolumn{2}{c}{Magnitude} & \multicolumn{3}{c}{Translation [cm]} & \multicolumn{3}{c|}{Rotation [\degree]} & Time [s] \\
        \textbf{count} & $E_\textrm{t}$ [cm] & $E_\textrm{R}$ [\degree] & x & y & z & roll & pitch & yaw \\
        \midrule
        \rowcolor{Gray}
        100 / 5.0\% & 2.94$\pm$0.13 & 0.14$\pm$0.00 & 0.66 & 2.78 & 0.49 & 0.03 & 0.05 & 0.13 & 3402 \\
        80 / 6.25\% & 3.32$\pm$0.65 & 0.14$\pm$0.01 & 0.58 & 3.24 & 0.31 & 0.01 & 0.05 & 0.13 & 3054 \\
        60 / 8.33\% & 3.46$\pm$0.88 & 0.14$\pm$0.02 & 0.21 & 3.41 & 0.34 & 0.02 & 0.04 & 0.13 & 3696 \\
        40 / 12.5\% & 3.38$\pm$0.41 & 0.14$\pm$0.02 & 0.38 & 3.32 & 0.41 & 0.02 & 0.05 & 0.13 & 3904 \\
        20 / 25.0\% & 3.96$\pm$0.49 & 0.18$\pm$0.02 & 0.89 & 3.80 & 0.41 & 0.02 & 0.06 & 0.16 & 4451 \\
        \bottomrule
    \end{tabular}
    \footnotesize
    This table corresponds to \cref{fig:ablation-pairs-correspondences}.
    For the motion constraints, we use 1,000 poses. We vary the pose count and the relative number of correspondences to obtain an approximately constant absolute number of correspondences.
\end{threeparttable}
% \vspace{-0.5cm}
\end{table}

\vfill\null

%%%%%%%%%%%%%%%%%%%%%%%%%%%%%%%%%%%%%%%%%%%%%%%%%%%%%%%%%%%%%%%%%%%%%%%%%%%%%%%%

\end{document}